\documentclass{article}

\usepackage{PRIMEarxiv}

\usepackage[utf8]{inputenc} 
\usepackage[T1]{fontenc}    
\usepackage{hyperref}       
\usepackage{url}            
\usepackage{booktabs}       
\usepackage{amsfonts}       
\usepackage{nicefrac}       
\usepackage{microtype}      
\usepackage{lipsum}
\usepackage{fancyhdr}       
\usepackage{graphicx}       
\graphicspath{{media/}}     
\usepackage{makecell}
\usepackage{booktabs}
\usepackage{pifont}
\usepackage{xcolor,colortbl}
\usepackage{multirow}
\usepackage{adjustbox}
\usepackage{tikz}
\usepackage{authblk}

\newcommand{\sourcetagmanual}[1]{\tikz[baseline=(X.base)]\node [draw=blue,fill=cyan!40,semithick,rectangle,inner sep=2pt, rounded corners=3pt] (X) {#1};}
\newcommand{\sourcetagautogen}[1]{\tikz[baseline=(X.base)]\node [draw=blue,fill=blue!30,semithick,rectangle,inner sep=2pt, rounded corners=3pt] (X) {#1};}
\newcommand{\sourcetagrulegen}[1]{\tikz[baseline=(X.base)]\node [draw=blue,fill=blue!20,semithick,rectangle,inner sep=2pt, rounded corners=3pt] (X) {#1};}
\newcommand{\sourcetagpub}[1]{\tikz[baseline=(X.base)]\node [draw=blue,fill=gray!40,semithick,rectangle,inner sep=2pt, rounded corners=3pt] (X) {#1};}

\newcommand{\evaltagmultichoice}[1]{\tikz[baseline=(X.base)]\node [draw=yellow,fill=pink!100!yellow!10,semithick,rectangle,inner sep=2pt, rounded corners=3pt] (X) {#1};}
\newcommand{\evaltaghumaneval}[1]{\tikz[baseline=(X.base)]\node [draw=yellow,fill=pink!40!red!20,semithick,rectangle,inner sep=2pt, rounded corners=3pt] (X) {#1};}
\newcommand{\evaltaggpteval}[1]{\tikz[baseline=(X.base)]\node [draw=yellow,fill=pink!40!blue!20,semithick,rectangle,inner sep=2pt, rounded corners=3pt] (X) {#1};}
\newcommand{\evaltagcaptionscore}[1]{\tikz[baseline=(X.base)]\node [draw=yellow,fill=pink!40!gray!20,semithick,rectangle,inner sep=2pt, rounded corners=3pt] (X) {#1};}

\newcommand{\containreasonstep}[1]{\tikz[baseline=(X.base)]\node [draw=red,fill=violet!30,semithick,rectangle,inner sep=2pt, rounded corners=3pt] (X) {#1};}

\NewDocumentCommand{\xudong}{ mO{} }{\textcolor{blue}{\textsuperscript{\textit{Xudong}}\textsf{\textbf{\small[#1]}}}}

\title{Exploring the Reasoning Abilities of Multimodal Large Language Models (MLLMs): A Comprehensive Survey on Emerging Trends in Multimodal Reasoning
}

\author{Yiqi Wang\thanks{Work done during an internship at ByteDance.}, Wentao Chen, Xiaotian Han, Xudong Lin, Haiteng Zhao\protect\footnotemark[1],
	Yongfei Liu,
	Bohan Zhai,
	Jianbo Yuan,\authorcr
	Quanzeng You\thanks{Corresponding author: \texttt{quanzeng.you@bytedance.com}.}, 
	Hongxia Yang}
\affil{ByteDance Inc.}

\begin{document}
\maketitle

\begin{abstract}
Strong Artificial Intelligence (Strong AI) or Artificial General Intelligence (AGI) with abstract reasoning ability is the goal of next-generation AI. 
Recent advancements in Large Language Models (LLMs), along with the emerging field of Multimodal Large Language Models (MLLMs), have demonstrated impressive capabilities across a wide range of multimodal tasks and applications.
Particularly, various MLLMs, each with distinct model architectures, training data, and training stages, have been evaluated across a broad range of MLLM benchmarks.
These studies have, to varying degrees, revealed different aspects of the current capabilities of MLLMs.
However, the reasoning abilities of MLLMs have not been systematically investigated.
In this survey, we comprehensively review the existing evaluation protocols of multimodal reasoning, categorize and illustrate the frontiers of MLLMs, introduce recent trends in applications of MLLMs on reasoning-intensive tasks, and finally discuss current practices and future directions. 
We believe our survey establishes a solid base and sheds light on this important topic, multimodal reasoning.
\end{abstract}


\section{Introduction}
In the past decade, with the help of increasing computational power and expanded datasets, Multimodal Large Language Models (MLLMs) have achieved remarkable progress in many domains and applications.
They are considered to be the most capable family towards the ultimate goal of Strong Artificial Intelligence (Strong AI)~\cite{searle1980minds} or Artificial General Intelligence (AGI)~\cite{goertzel2007artificial}. 
Strong AI is often thought to possess a mind, the question of whether Multimodal Large Language Models have minds, or how one might determine their existence, remains an open and complex topic.
One does not need to have Sherlock Holmes' mind to reason over observations naturally in multiple modalities from the real world, such as vision, audio, text, smell, \textit{etc.}, and subsequently take action. 
In the Dual-system theory~\cite{evans2003two,evans2013dual}, a widespread theory of cognitive science, the second type of the human cognition systems, which enable abstract reasoning, is considered to be ``\textit{evolutionarily recent and distinctively human}''.
This characterization prompts an intriguing inquiry: are MLLMs capable of reasoning?

Specifically, we are interested in the reasoning tasks that require the comprehension and integration of information from various modalities, including vision, text, audio, and others. 
MLLMs have demonstrated effectiveness in a variety of multimodal reasoning tasks.
Notable examples include Visual Question Answering (VQA)~\cite{balanced_vqa_v2,marino2019ok,saikh2022scienceqa,hudson2019gqa}, multimodal dialogue~\cite{bai2023touchstone,li2023scigraphqa,huang2023sparkles}, among others. 
Recently, there has also been extensive research focusing on particularly improving the reasoning abilities of MLLMs, such as multimodal instruction tuning~\cite{alayrac2022flamingo,liu2023llava,li2023empowering,huang2023sparkles} and multimodal reasoning by prompting~\cite{tsimpoukelli2021multimodal,zeng2022socratic}. 
The strong capabilities of MLLMs also intrigue research on embodying them as agents in real-world environment~\cite{brohan2023rt,chen2023pali,driess2023palme} or enabling them using tools~\cite{wu2023visual,yang2023mmreact}.
Despite the impressive performance on existing benchmark datasets~\cite{alayrac2022flamingo,li2023blip2,liu2023llava,zhang2023llamaadapter,driess2023palme,peng2023kosmos2,ye2023mplugowl}, it is still too early to conclude that current MLLMs can truly reason over multiple input modalities. 

Even in the text-only setting, Large Language Models (LLMs) are still known to lack proper reasoning abilities in some aspects, such as mathematical~\cite{frieder2023mathematical}, and multi-hop reasoning~\cite{bang2023multitask,ji2023survey}. 
Moreover, both LLMs and MLLMs are also known to have hallucination issues~\cite{manakul2023selfcheckgpt}, which prevent the models from reasoning properly.
Motivated by the significant importance of reasoning abilities in MLLMs and the fast-paced development of related research, we believe it is necessary to comprehensively review the current status of the reasoning abilities of MLLMs and provide insightful discussion to illuminate future research. 
In the following sections, we will initially define the reasoning abilities of interest and illustrate existing evaluation protocols (Sec. \ref{sec:nlr}); subsequently, we will summarize the current status of MLLMs (Sec. \ref{sec:mllm}); this will be followed by an in-depth look at a key training phase for multimodal reasoning, namely, multimodal instruction tuning  (Sec. \ref{sec:mit}); 
then we will discuss the reasoning-intensive applications of MLLMs, including Embodied AI (Sec. \ref{sec:emboded}) and tool usage (Sec. \ref{sec:tool}); 
afterward, we will analyze results on multimodal reasoning benchmarks (Sec. \ref{sec:eval}). 
Finally, we will provide concentrated insights and discussions on the current status and future directions of MLLMs (Sec. \ref{sec:conclusion}).
\section{Reasoning: Definition and Evaluation Protocols}
\label{sec:nlr}

\subsection{What is Reasoning?}
Reasoning is one of the fundamental intelligent behaviors of human beings, which requires understanding and analyzing given conditions and background knowledge to derive a new conclusion logically and rationally \cite{yu2023nature, huang2022towards, walton1990reasoning}. 
Reasoning has been extensively studied within the field of logic~\cite{delancey2017concise,finocchiaro1984informal,madden1994reasoning}. 
To obtain a clear understanding of reasoning, we refer to the definitions established in the field of logic~\cite{bronkhorst2020logical,dowden2018logical}, which defines reasoning based on the concepts of premises, conclusions, and inferences. 
Reasoning is usually regarded as the integration of these concepts. 
To be more specific, premises and conclusions are true or false claims about a case. 
Inferences are the intermediate reasoning steps that select and interpret information from given premises, make connections, verify, and ultimately draw conclusions based on the provided and interpreted data.

Reasoning in the field of logic highly relies on mathematics \cite{madden1994reasoning},  which is used to construct a set of basic logic rules \cite{rybakov1997admissibility}. 
Accordingly, only reasoning that adheres to these logical rules is considered valid.
Apart from logic rules, domain knowledge is also required to perform practical reasoning tasks \cite{yang2006belief}. 
For instance, arithmetic reasoning necessitates mathematical knowledge, while commonsense knowledge is essential for reasoning in daily life tasks. 
The domain knowledge serves as additional premises besides the given inputs and is indispensable to obtaining valid conclusions in specific fields.

According to \cite{bronkhorst2020logical}, reasoning can be divided into formal and informal reasoning, where conclusions of formal reasoning are guaranteed to be true as long as the premises are true, while informal reasoning does not guarantee the truth of conclusions especially when the available information is incomplete or ambiguous. 
In usual, informal reasoning is performed with natural language and is essential for daily-life tasks. 
Additionally, based on the direction of inference, reasoning can be divided into deductive, inductive, abductive, and analogical reasoning:

\begin{itemize}
    \item \textbf{Deductive reasoning} \cite{johnson1999deductive} represents the most classical form of reasoning. 
    Given a set of known knowledge (premises), it infers new knowledge step by step to obtain the conclusion. 
    For example, given the premises that ``cats are mammals'' and ``all mammals have four feet'', deductive reasoning can infer a new conclusion that ``cats have four feet''. 
    Note that deductive reasoning only concerns that the inference steps follow the logic rules and does not impose any restrictions on the truth of premises. 
    Therefore, wrong premises may lead to wrong conclusions even if the reasoning steps are logical.
    \item \textbf{Inductive reasoning} concentrates on inferring general rules from specific observations \cite{li1992inductive}. 
    For example, given the premises (observations) that ``any mammal that I have seen so far has four feet'', induction reasoning can infer that ``all mammals have four feet". 
    Inductive reasoning is an effective tool in scientific areas to discover new principles and laws. 
    Note that since it is hard to collect complete observations, the conclusions of inductive reasoning may be incorrect for some unseen observations.
    \item \textbf{Abductive reasoning} is to infer the best explanation for the given observations \cite{douven2011abduction}. 
    It is perceived as the backward direction of deductive reasoning, where multiple reasons can lead to the results (observations) and the most likely reason should be inferred. 
    Consider this scenario: a car is parked on the highway with its hazard lights flashing.
    Abductive reasoning could lead to the more plausible conclusion that the car is broken down, rather than the less likely explanation of someone playing a prank. 
    Since the number of possible reasons is usually massive, abduction reasoning requires a lot of commonsense and domain knowledge to infer a credible reason.
    \item \textbf{Analogical reasoning} involves transferring knowledge from one or several instances to another, based on their similarities \cite{goswami1991analogical}. 
    Two forms of analogical reasoning were studied and applied in real-life activities \cite{bartha2013analogy}. 
    The first form takes as inputs one or multiple similar cases, then arrives at a hidden proposition, and finally applies the proposition to a new case. 
    For example, consider the two cases that ``iron can conduct electricity'' and ``copper can conduct electricity".
    From this, one might infer a proposition that ``any metal can conduct electricity'', thereby inferring that ``silver, being a metal, can also conduct electricity''. 
    The second form of analogical reasoning considers the similarity of two entities to infer a property in one based on the property of the other. 
    For instance, given the premises that ``exposing plants to adequate sunlight enhances their growth and health'' and ``humans and plants both require certain environmental factors to thrive, like water, air, and nutrients'', one could use analogical reasoning to hypothesize that ``humans may also benefit from regular exposure to sunlight for their health and well-being'' \cite{salmon2012introduction}. 
    With analogical reasoning, the properties of a new object can be quickly inferred at a low cost. 
    However, the premises of analogical reasoning can only support the conclusions that are likely rather than definitively true.
\end{itemize}

In this paper, we focus on the reasoning abilities of Multimodal Large Language Models. 
The reasoning methods employed by these models fall under the category of informal reasoning. 
This is primarily because they utilize natural language to articulate the steps and conclusions involved in the reasoning process and they allow a certain degree of inaccuracy in their reasoning mechanisms.
This paper primarily focuses on three reasoning types: \textbf{deductive reasoning}, \textbf{abductive reasoning}, and \textbf{analogical reasoning}.
These types are highlighted due to their prevalent application in real-world reasoning tasks, particularly within the scope of current MLLMs.

\subsection{Language-only Reasoning Tasks}
To gain a deeper insight into the reasoning abilities of Multimodal Large Language Models, it is crucial to understand the associated reasoning tasks. 
These tasks are widely regarded as requiring the models' reasoning capabilities for effective resolution. 
We can divide reasoning tasks into two categories based on the input data: language-only reasoning tasks and multimodal reasoning tasks, with the former requiring no images and the latter involving both images and text. 
Research on language-only tasks has a longer history, and the methodologies used for task categorization and the insights gained from these studies offer valuable guidance for the development of multimodal reasoning tasks. 
In this section, we will present four distinct types of language-only reasoning tasks: solving math problems, engaging in commonsense reasoning, tackling symbolic reasoning tasks, and interacting with various environments.

\subsubsection{Solving Math Problems}

Solving math problems usually requires one-step or multistep arithmetic reasoning. 
Based on the understanding of input question, the implicit arithmetic operations, and concept knowledge, a solver should infer a sequence of operation steps that can derive final answers.
The range of implicit operations and conceptual knowledge can be categorized according to different school grade levels.
For example, the benchmarks of GSM8K \cite{cobbe2021training}, SVAMP \cite{patel2021nlp}, ASDiv \cite{miao-etal-2020-diverse}, and MAWPS \cite{koncel-kedziorski-etal-2016-mawps} require mathematical knowledge typically acquired in primary schools. 
This includes fundamental operations such as addition, subtraction, multiplication, and division. 
The MathQA \cite{amini2019mathqa} benchmark and the AQuA \cite{ling2017program} benchmark include mathematical problems sourced from standardized tests such as the GMAT (Graduate Management Admission Test) and the GRE (General Test). 
The MATH \cite{hendrycks2021measuring} benchmark features highly challenging mathematical problems, including areas such as permutation and combination problems, geometric series problems, solving high-order equations, among others. 
This benchmark requires solvers to have a lot of advanced mathematical knowledge and mathematical reasoning techniques, as well as the ability to follow a multi-step procedure, and thus remains a very challenging task. 

\subsubsection{Engaging in Commonsense Reasoning}

Commonsense is a broadly encompassing yet somewhat loosely defined concept. 
While lacking exact boundaries, this generally refers to knowledge that goes beyond specialized expertise and is expected to be well-known to individuals who have completed basic education. 
Commonsense knowledge extends across various domains, including social common sense (\textit{e.g.}, understanding that people would feel embarrassed if publicly blamed), physical common sense (\textit{e.g.}, recognizing that a car is faster than a bicycle), biological common sense (\textit{e.g.}, knowing that penguins and koalas do not naturally encounter each other), and numerous other areas. 
It is widely recognized that commonsense knowledge plays a significant role in everyday decision-making and real-life scenarios, making commonsense reasoning a fundamental prerequisite for language models. 
Recent studies have introduced several datasets aimed at commonsense reasoning, such as HellaSwag \cite{zellers2019hellaswag}, Winogrande \cite{sakaguchi2020winogrande}, Socialiqa \cite{sap2019social}, Piqa \cite{bisk2020piqa}, CommonGen \cite{lin2020commongen}, Cosmos QA \cite{huang2019cosmos}, and ART \cite{bhagavatula2019abductive}, which usually do not require a multi-step reasoning process.

\subsubsection{Tackling Symbolic Reasoning}

Symbolic reasoning can be characterized as a cognitive process carried out on abstract objects, guided by precisely defined rules like logical derivation. 
Besides coding and mathematical problem-solving, there exist various tasks that necessitate the application of symbolic reasoning.
One such task is logistic reasoning, exemplified by datasets like PrOntoQA \cite{saparov2022language}, SimpleLogic \cite{zhang2022paradox}, FOLIO \cite{han2022folio}, and ProofWriter \cite{tafjord2021proofwriter}. 
In these tasks, a set of facts and logical rules are given in the context and the model is required to prove a formula based on logical operations. 
Other tasks involve the comprehension of virtual objects. 
For instance, in datasets such as Penguins, Date,  and Colored Objects from BIG-Bench Hard \cite{suzgun2022challenging}, there is a demand for statistical analysis and manipulation of properties related to virtual objects.
An illustrative question might be, "Which penguin is named after a famous jazz musician?". 
While language models demonstrates the ability to understand simple symbolic manipulations, they have been identified as less adept in complex symbolic reasoning tasks \cite{saparov2022language}.

\subsubsection{Interacting with Various Environments}
Interacting with environments require reasoning skills that involve using commonsense knowledge to understand the current situation and plan future actions. 
Additionally, these environments demand the ability to process feedback and adjust subsequent actions based on that feedback.
This domain has seen extensive exploration in recent studies focusing on language model-driven agents. 
Many of these environments primarily employ textual modalities as their basis.
Examples include Textworld \cite{cote2019textworld}, Alfworld \cite{shridhar2020alfworld}, and WebShop \cite{yao2022webshop}. 
These virtual scenarios rely on textual descriptions for state representation, feedback, and interaction. 
In these environments, agents are tasked with completing various objectives. 
For instance, they might be required to place fresh lettuce on a dining table within a living room setting or to choose a 3-ounce bottle of citrus-scented deodorant from a simulated online shopping platform. 
Beyond purely textual environments, some environments incorporate multiple modalities to challenge language model reasoning. 
For instance, BabyAI \cite{chevalier2018babyai} and MineDojo \cite{fan2022minedojo} present game-like scenarios. 
Within these environments, models are expected to engage in tasks such as crafting an iron sword in the Minecraft game. 
Furthermore, a subset of studies has delved into the realm of reasoning within real-world contexts. 
For example, Ahn \textit{et al.} \cite{ahn2022can} tackled 101 robotic tasks within an actual kitchen, pushing the boundaries of language model reasoning beyond simulated scenarios.

\subsection{Multimodal Reasoning Benchmarks}
\label{sec:mm_bench}

With the ever-evolving multi-modal large language models' reasoning capability, benchmarks play a pivotal role in assessing their performance, understanding their capabilities, and identifying areas that demand improvement. 
Based on the aforementioned definition and categorization of reasoning, an ideal multimodal reasoning benchmark is supposed to 1) truly require multimodal information; 2) follow the categorization of reasoning; and 3) have detailed annotation of reasoning steps to examine whether the reasoning process is correct.

\subsubsection{Benchmark Datasets}

Traditional datasets, such as COCO caption \cite{lin2014microsoft}, Nocaps \cite{agrawal2019nocaps}, and Flickr30K \cite{plummer2015flickr30k}, are utilized to evaluate the MLLMs' comprehension of image content and captioning. 
Datasets focused on visual question answering, such as VQAv2 \cite{balanced_vqa_v2}, OK-VQA \cite{marino2019ok}, ScienceQA \cite{saikh2022scienceqa}, and GQA \cite{hudson2019gqa}, feature question-answer pairings. 
These datasets serve as an initial platform for evaluating a model's reasoning capabilities. 
Recently, more evaluation benchmarks have been introduced to holistically compare the various capabilities of MLLMs. 
We summarize existing evaluation benchmarks in \tablename~\ref{tab:benchmark_tab}, with an emphasis on their reasoning-related evaluation.

Surprisingly, except for the recently introduced Infi-MM-Eval~\cite{han2023infimmeval}, most existing reasoning-related multimodal benchmarks are not reasoning-focused and do not follow the aforementioned criteria of an ideal benchmark. 
For example, another manually labeled benchmark MM-Vet~\cite{yu2023mmvet} does not follow the reasoning categorization to properly evaluate different reasoning types, nor contain reasoning steps to examine the reasoning process. 
Another example is MMMU~\cite{yue2023mmmu},  a dataset collected from textbooks.
It does not specify the type of reasoning required and only includes partial annotations of the reasoning steps. 
For simplicity and to ensure our empirical findings are more focused on reasoning and robustness, our analysis will primarily concentrate on these three datasets in Sec. \ref{sec:eval}.

\subsubsection{Evaluation Metrics}

With these rich evaluation datasets shown in \tablename~\ref{tab:benchmark_tab}, it is important to evaluate using proper metrics. Therefore, it is necessary to categorize and review the evaluation metrics. 
The complexity of evaluating MLLMs arises from the multimodal nature of these tasks. 
The metrics must capture the synergistic relationship between language and vision, as well as other possibly involved modalities. 
This relationship demands assessments that not only estimate the accuracy of predictions but also the depth, nuance, and relevancy of multimodal associations. 

\clearpage

\begin{table}[!htbp]
    \tiny
    \caption{Evaluation Benchmark Summarization. Benchmarks are ordered by releasing date.}
    \centering
    \begin{tabular}{l|p{3.5cm}|p{2cm}|p{2cm}|p{6cm}}
    \toprule
    Dataset & Description & Stats & Reasoning Part & Tags \\
    \midrule
    MMMU\cite{yue2023mmmu} & Designed to evaluate multimodal models on massive multi-discipline tasks demanding college-level subject knowledge and deliberate reasoning & 11.5K questions from college exams, quizzes, and textbooks & & \sourcetagmanual{Manual Collect}, \evaltagmultichoice{Multiple-Choice Accuracy} \\
    \midrule
    InfiMM-Eval\cite{han2023infimmeval} & Designed for open-ended complex reasoning QA with intermediate steps & 279 high quality samples &  & \sourcetagmanual{Manual Collect}, \evaltagmultichoice{GPT eval}, \containreasonstep{Full reasoning step} \\
    \midrule
    HallusionBench\cite{guan2023hallusionbench} & benchmark for evaluating both visual and language hallucination &  346 images with 1129 questions & & \sourcetagmanual{Manual Collect}, \sourcetagpub{Public Source}, \evaltagmultichoice{GPT eval} \\
    \midrule
    MathVista\cite{lu2023mathvista} & Consolidated Mathematical reasoning benchmark with visual contexts, covers 7 types of reasoning, \textit{e.g.} algebraic, arithmetic, geometry, logical, numeric common sense and scientific & 6,141 samples & & \sourcetagmanual{Manual Collect}, \sourcetagpub{Public Source}, \evaltagmultichoice{Multiple-Choice Accuracy} \\
    \midrule
    LLMDoc\cite{ye2023mplugdocowl} & OCR-free document instruction understanding evaluation set, including table, chart, document, natural image, and webpage & 100 samples & & \sourcetagmanual{Manual Collect}, \sourcetagpub{Public Source}, \evaltaghumaneval{Human eval} \\
    \midrule
    PCA-Eval\cite{chen2023endtoend} & Evaluating the decision-making ability of embodied agents from three perspectives: perception, cognition, and action. Images are from various embodies environments & 300 multi-choice questions & each question has a "reason" field for evaluating reasoning capability & \sourcetagmanual{Manual Collect}, \evaltagmultichoice{Multiple-Choice Accuracy}\\
    \midrule
    SparklesEval\cite{huang2023sparkles} & GPT-assisted benchmark for assessing a model’s conversational competence across multiple images and dialogue turns & 150 dialogs, total 550 images & & \sourcetagautogen{GPT Generate}, \evaltaggpteval{GPT eval} \\
    \midrule
    M3Exam\cite{zhang2023m3exam} & Multilingual, multimodal, multilevel, multiple choice questions from exam papers & 12137 questions, 23\% contain images & & \sourcetagmanual{Manual Collect}, \evaltagmultichoice{Multiple-Choice Accuracy} \\
    \midrule
    TouchStone\cite{bai2023touchstone} & Comprehensive visual dialogue dataset, consisting of open-world images and questions, covering 5 categories of abilities and 27 subtasks & 908 questions & comprehension questions at 29.6\%, 3.6\% for multi-image analysis & \sourcetagmanual{Manual Collect}, \evaltaggpteval{GPT eval} \\
    \midrule
    MM-Vet\cite{yu2023mmvet} & QAs for 6 VL capabilities with 187 online images and 13 public dataset images & 218 questions & 11.9\% of math and 34.4\% of spatial awareness & \sourcetagmanual{Manual Collect}, \evaltaggpteval{GPT eval} \\
    \midrule
    MMBench\cite{liu2023mmbench} & Vision-language QAs in the format of multiple choice over 20 abilities & 2,974 questions & 1114 Reasoning Questions & \sourcetagmanual{Manual Collect}, \evaltagmultichoice{Multiple-Choice Accuracy}\\
    \midrule
    MME\cite{fu2023mme} & Yes and No choice questions on 14 subtasks. Each image has 2 questions & 1097 images & 130 images & \sourcetagmanual{Manual Collect}, \evaltagmultichoice{Multiple-Choice Accuracy} \\
    \midrule
    LAMM-Benchmark\cite{yin2023lamm} & 9 image tasks and 3 point cloud tasks & 62439 images and 12788 point cloud samples, 186k instruction-response pairs & ScanQA, SQAimage and AI2D & \sourcetagautogen{GPT Generate}, \evaltagmultichoice{Traditional Metrics}, \evaltaggpteval{GPT eval} \\
    \midrule
    CCEval\cite{zhai2023halleswitch} & benchmark for evaluating detailed caption hallucination with object coverage & 100 VisualGenome images & & \sourcetagpub{Public Source}, \evaltagmultichoice{GPT eval}\\
    \midrule
    VisIT-Bench\cite{bitton2023visitbench} & Instruction following evaluation with human-authored instruction-conditioned caption & 592 queries with 1159 images & & \sourcetagmanual{Manual Collect}, \evaltaggpteval{GPT eval}, \evaltaghumaneval{Human eval} \\
    \midrule
    SEED-Bench\cite{li2023seedbench} & Multiple choice questions, spans 12 evaluation dimensions in images and videos. Questions generated by GPT, answered by human & 19K questions & 331 visual reasoning, 97 instance interaction and 657 spatial relations & \sourcetagmanual{Manual Collect}, \sourcetagautogen{GPT Generate}, \evaltagmultichoice{Multiple-Choice Accuracy} \\
    \midrule
    I4\cite{li2023empowering} & Evaluating instruction following ability on interleaved VL instructions, covers 19 scenarios with 29 tasks & 18k instructions with 62k images & & \sourcetagpub{Public Source}, \evaltagmultichoice{Multiple-Choice Accuracy}, \evaltagcaptionscore{Caption Score} \\
    \midrule
    LVLM-eHub\cite{shao2023tiny} & Combination of many public datasets and an online arena platform & 47 visual benchmarks & DocVQA, OKVQA, ScienceQA, SNLI-VE \textit{etc.} & \sourcetagpub{Public Source}, \evaltaghumaneval{Human eval}, \evaltagcaptionscore{Caption Score}, \evaltagmultichoice{Multiple-Choice Accuracy} \\
    \midrule
    OwlEval\cite{ye2023mplugowl} & 50 collected images with questions, some has multi-turn conversations & 82 questions & & \sourcetagmanual{Manual Collect}, \evaltaghumaneval{Human eval} \\
    \midrule
    Open-VQA\cite{zeng2023matters} & Open-ended questions including images and videos & 450 questions & 31 Reasoning Questions & \sourcetagmanual{Manual Collect}, \evaltaghumaneval{Human eval} \\
    \midrule
    ScienceQA\cite{lu2022learn} & Multimodal multiple-choice questions with annotations and explanations covers natural science, social science, and language science, collected from online education platform & 21208 questions, 10332 contain images & & \sourcetagmanual{Manual Collect}, \evaltagmultichoice{Multiple-Choice Accuracy} \\
    \midrule
    Winoground\cite{thrush2022winoground} & Measuring visio-linguistic compositional reasoning & 1600 image-text pairs, comprise 400 examples & & \sourcetagmanual{Manual Collect}, \evaltagmultichoice{Multiple-Choice Accuracy} \\
    \midrule
    RAVEN\cite{zhang2019raven} & Multi-choices Raven’s Progressive Matrices (RPM) questions & 1,120,000 images and 70, 000 RPM problems & & \sourcetagrulegen{Rule Generate}, \evaltagmultichoice{Multiple-Choice Accuracy} \\
    \bottomrule
    \end{tabular}
    \label{tab:benchmark_tab}
\end{table}

\clearpage
The existing evaluation metrics can be categorized from different aspects. 
Although we mainly focus on three datasets for results analysis, we categorize all the related datasets for a comprehensive summarization.
From the perspective of evaluating the different capabilities of models, evaluation can be categorized into 3 types:

\begin{itemize}
    \item For \textbf{\textit{vision capability}}, traditional perception metrics are relevant, including tasks such as identifying, classifying, and locating objects within an image, as well as interpreting those objects in the context of the overall scene.
    The common metrics used are image classification accuracy (ImageNet \cite{5206848}), object detection mAP (COCO \cite{lin2015microsoft}), object segmentation mIoU (LVIS \cite{gupta2019lvis}), and others.
    \item For \textbf{\textit{linguistic capability}}, commonsense understanding and coherence needed to be considered as metrics. 
    It goes beyond merely generating grammatically correct sentences; it's about producing responses that are contextually relevant, logical, and semantically coherent, ensuring alignment with the given visual input. 
    The common metrics used are BLEU \cite{papineni-etal-2002-bleu}, CIDEr \cite{vedantam2015cider}, ROUGE \cite{lin-2004-rouge}, among others.
    \item For \textbf{\textit{reasoning capability}}, it covers spatial reasoning, knowledge reasoning (including areas like common sense, mathematics, text, code, and more), as well as hypothesis-based reasoning. 
    Just as humans deduce information based on textual and visual cues, MLLMs should exhibit the ability to reason, draw inferences, and predict outcomes based on the multi-modal data they process. 
    The common metrics are QA-Accuracy \cite{yue2023mmmu}, Elo score \cite{zheng2023judging}, GPT-4 \cite{openai2023gpt4} evaluation, and others.
\end{itemize}

From the types of evaluation questions, existing multimodal benchmarks can be divided into two distinct categories:
\begin{itemize}
    \item \textbf{\textit{Closed-Set Evaluation Benchmarks}} are circumscribed by a predetermined set of categories or outcomes. 
    When MLLMs are assessed under this paradigm, they're essentially provided a limited set of potential responses or classifications. 
    Such evaluations gauge the model's aptitude to correctly categorize or respond within the given confines, making them particularly suited for tasks where boundaries are well-defined and the space of responses is finite. 
    Certain visual evaluation tasks are of a closed-set nature, such as image classification and attribute classification.
    Conversely, some benchmarks with open-ended questions force to reformat answers into multiple-choice  format, examples of which includes MMBench~\cite{liu2023mmbench}, MME~\cite{fu2023mme} and ScienceQA~\cite{lu2022learn}. 
    However, as studied in a recent work \cite{zong2023fool}, authors found that popular MLLMs are vulnerable to adversarial permutation in answer sets for multiple-choice prompting.
    \item \textbf{\textit{Open-Set Evaluation Benchmarks}} are more unbounded and exploratory. 
    They allow models to generate responses without confining them to a predefined set.
    This evaluates the model's capability to venture into the unknown, generating or deducing outcomes that might not be part of the training set. 
    In the vast and dynamic domain of multimodal interactions, where real-world scenarios often elude strict categorizations, open-set evaluations become crucial in assessing a model's generalization capabilities and adaptiveness to novel inputs. 
    Two common approaches for evaluating open-set answers are human scoring such as LVLM-eHub~\cite{shao2023tiny} and Lynx~\cite{zeng2023matters}, and automatic language model scoring, like TouchStone~\cite{bai2023touchstone} and VisIT-Bench~\cite{bitton2023visitbench}. 
\end{itemize}

With the three datasets subject to our analysis, diverse reasoning evaluation metrics are employed, including GPT-4 evaluation (Infi-MM-Eval, MM-Vet) and QA-Accuracy (MMMU).
Additionally, these datasets feature different evaluation question setups: Open-Set (Infi-MM-Eval, MM-Vet) and Closed-Set (MMMU). 
Therefore, a representative subset of benchmarks are covered in our analysis.

\subsection{Improving Reasoning Abilities for LLMs}

Since reasoning abilities are essential for solving complex tasks, a wealth of research has focused on improving reasoning abilities in LLMs. 
Using supervised data to pretrain or fine-tune LLMs is a straightforward method to enhance their reasoning capabilities. 
For instance, during the pretraining stage, Lewkowycz \textit{et al}. \cite{lewkowycz2022solving} incorporate a variety of mathematical corpora into their training set to improve quantitative reasoning abilities. 
In the fine-tuning stage, Rajani \textit{et al}. \cite{rajani2019explain} utilize a small instruction-tuning dataset to fine-tune the pretrained GPT model. 
Beyond supervised training, in-context learning \cite{wei2022chain,zhou2022least} and prompt engineering \cite{kojima2022large} have also shown great potential for improving reasoning abilities. such as the Chain-of-Thought approach \cite{wei2022chain} and its variants \cite{fu2022complexity,zhang2022automatic}. 
Another line of research \cite{gao2023pal,chen2022program,yao2023react} has proposed interacting LLMs with external tools, such as a Python code interpreter \cite{yao2023react}, to enhance the accuracy of each reasoning step. 
Within the domain of MLLMs, these methods, namely instruction-tuning, prompt engineering, and tool usage have also been explored for multimodal reasoning, which we will later elaborate.
\section{Multimodal Large Language Models}
\label{sec:mllm}
\subsection{Overview}
Multimodal Large Language Models typically comprise a visual encoder and a language model, linked by a connector. 
Both the visual encoder and the language model are commonly initialized from pretrained models. 
The distinctions among MLLMs can be analyzed based on the following key aspects:

\textbf{Training Data:} Multimodal training data can be segmented into three parts.  \textbf{(1)} Pretraining data is utilized for aligning other modalities, such as the visual modality, with the Large Language Model.
\textbf{(2)} (Optional) High-quality multi-task supervised learning data comprises question-answer pairs used to inject knowledge into the model.
\textbf{(3)} The instruction tuning data enhances Multimodal Large Language Models by improving their abilities to follow instructions and engage effectively in user interactions. 

\begin{figure}[htp]
    \centering
    \includegraphics[width=1.0\textwidth]{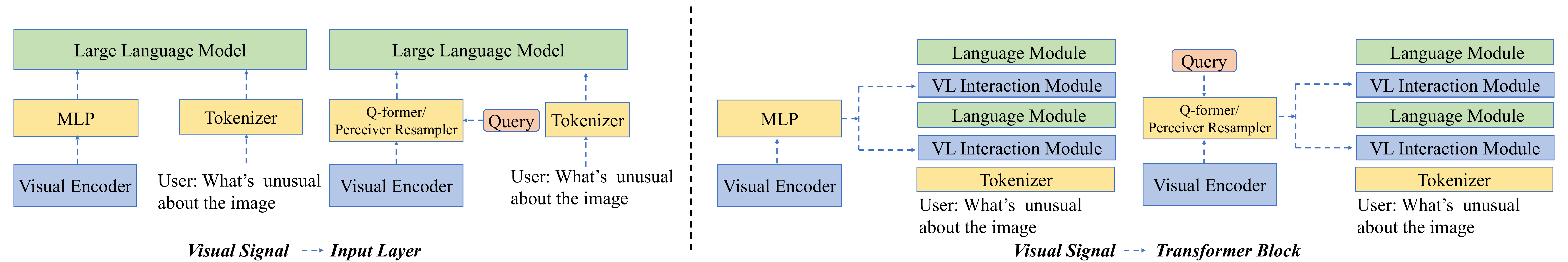}
    \caption{Various architectures of Multimodal Large Language Models.}
    \label{fig:mllm:vision_signal}
\end{figure}

\textbf{V-L Interaction Module:} As shown in Figure~\ref{fig:mllm:vision_signal}, some models incorporate visual tokens as if they were a foreign language, directly injecting them into the input layer.
For instance, LLaVA~\cite{liu2023llava} directly inputs the visual signal into the input layer of the large language model.
In contrast, other models, such as Flamingo~\cite{alayrac2022flamingo}, employ cross-attention layers to facilitate interactions between visual and language features within transformer blocks. 

\textbf{Connector Architecture:} The specifics of the connector design also play a pivotal role in determining the capabilities of MLLMs. 
A key differentiating factor among these models is their choice of visual-language connectors. 
Models such as BLIP-2~\cite{li2023blip2}, Flamingo~\cite{alayrac2022flamingo}, and QWen-VL~\cite{bai2023qwenvl} utilize query-based connectors like Q-former/perceiver resampler. 
Conversely, LLaVA~\cite{liu2023llava} and MiniGPT4-v2~\cite{zhu2023minigpt} employ a multilayer perceptron (MLP) as the connector.

\subsection{Large Language Models}
Recent advancements in MLLMs represent an evolution of LLMs, which makes reviewing of LLM development both relevant and beneficial.
GPT-3 \cite{NEURIPS2020_1457c0d6} pioneered this domain, showcasing a powerful emergent capability in zero-shot generation tasks. 
This was evident in its in-context learning and chain-of-thought processes, achieved using 175B parameters and pretraining on extensive web-text data. 
Subsequently, ChatGPT \cite{ouyang2022training} further augmented these capabilities by incorporating instruction tuning and reinforcement learning from human feedback. 
GPT-4 is an advanced iteration of ChatGPT; however, its training specifics remain undisclosed due to the source code not being made public. 
Other noteworthy LLMs, such as PaLM \cite{chowdhery2022palm}, BLOOM \cite{workshop2023bloom}, Chinchilla \cite{hoffmann2022training}, LLaMA \cite{touvron2023llama}, OPT \cite{zhang2022opt},  GLM \cite{du-etal-2022-glm}, Alpaca \cite{alpaca} and Vicuna \cite{vicuna2023}, have also been introduced, propelling the domain of LLMs forward.

\subsection{Multimodal Large Language Models}
\begin{figure}[htp]
    \centering
    \includegraphics[width=0.9\textwidth]{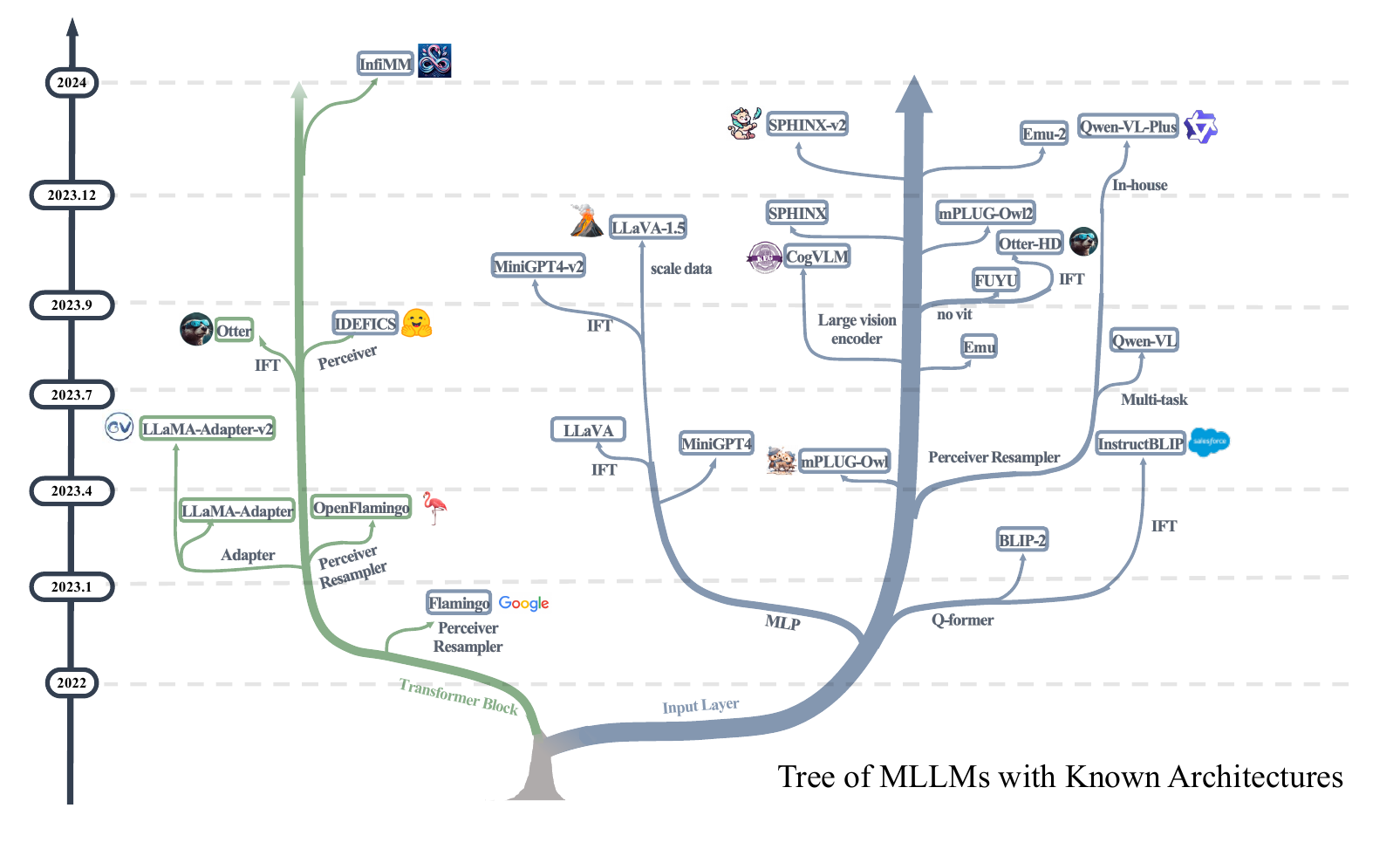}
    \caption{A brief timeline outlining recent developments in MLLMs.} 
    \label{fig:mllm:tree}
\end{figure}

 We give a brief timeline in Figure~\ref{fig:mllm:tree}, the efficacy of Multimodal Large Language Models is enhanced by incorporating vision encoders and connectors.
 This integration allows for training on diverse multimodal datasets. 
 \tablename~\ref{tab:multimodal_tab} highlights key distinctions among various MLLMs. 
 The Flamingo architecture \cite{alayrac2022flamingo} pioneers the usage of query-based cross-attention mechanism, termed the ``perceiver resampler''. 
 This innovative approach constructs a robust vision-language interactive module, showcasing remarkable abilities in in-context learning.
 BLIP-2~\cite{li2023blip2}, integrates the Q-Former.
 This query-based sampler, initialized from pretrained BERT, aims to bridge the gap between vision and language modalities.

These methods demonstrate their proficiency in generating text from images, a capability achieved through various stages of pretraining alignment. 
Later, InstructBLIP \cite{dai2023instructblip}, enhances the pretrained BLIP-2 by training on instruction tuning datasets sourced from diverse public datasets. 
This augmentation results in improvements across the majority of zero-shot performance metrics, surpassing the performance of BLIP-2. 
LLaVA~\cite{liu2023llava}  further enhances the instruction-following ability of MLLMs by training on visual instruction tuning data. 
Its subsequent iteration, \textit{LLaVA-1.5} \cite{liu2023improved}, expands upon this by integrating VQA datasets into the instruction tuning data, leading to significant performance improvements across a range of benchmarks.  
Notably, both InstructBLIP and LLaVA-1.5 follow a two-stage training process, which includes pre-training and subsequent instruction tuning. 

In contrast, QWen-VL \cite{bai2023qwenvl} introduces a more complex three-stage training process. 
This process includes pretraining, multi-task training, and instruction tuning. 
A particular emphasis is placed on the use of high-quality supervised data during the multi-task training stage.
Additionally, QWen-VL attempts to fine-tune all modules of the model, including the vision encoder, connector, and the large language model.

Furthermore, MiniGPT-v2~\cite{zhu2023minigpt} also employs a three-stage framework.
This framework emphasizes task-specific instruction templates for diverse tasks and involves fine-tuning both the projection layer and the Large Language Model, showcasing considerable improvements over MiniGPT4. 
LLaMA-Adapter \cite{zhang2023llamaadapter} incorporates a lightweight adaptation method for fine-tuning. 
This method excels in instruction-following and multimodal reasoning tasks, particularly image-based question answering. 
Additionally, mPLUG-owl2 \cite{ye2023mplugowl} leverages modality collaboration.
This approach enhances performance in both text-only and multimodal tasks. 
Meanwhile, Otter \cite{li2023otter} refines the OpenFlamingo model, which focuses on improving adherence to instructions and effectively utilizing in-context samples. 
Notably, the CogVLM~\cite{wang2023cogvlm} demonstrates superior performance by integrating an extra visual expert in the language model, achieving state-of-the-art performance on several benchmarks.

\begin{figure}[htp]
    \centering
    \includegraphics[width=0.9\textwidth]{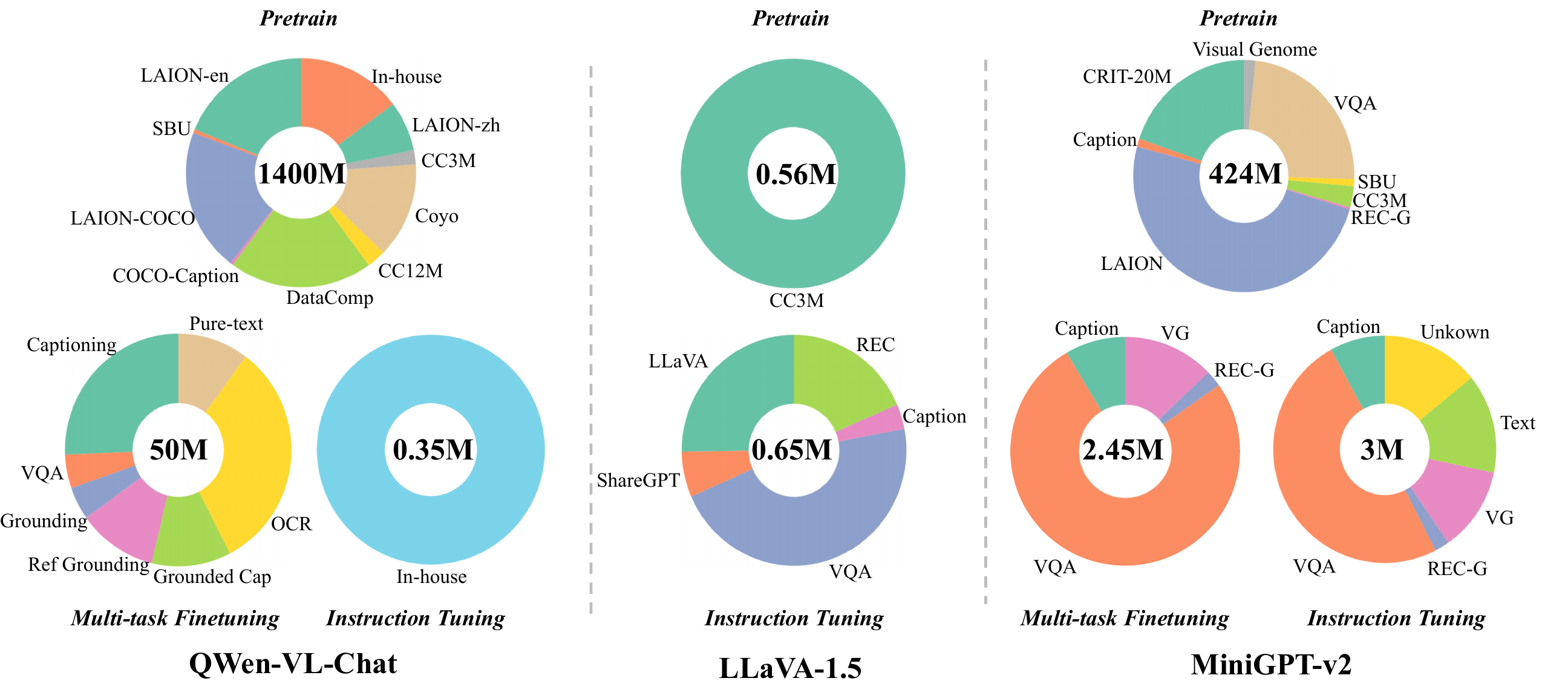}
    \caption{The datasets used in different stages of various Multimodal Large Language Models are depicted. At the center of each circle, the total number of samples (in millions) for each stage is displayed.}
    \label{fig:mllm:MLLM_example}
\end{figure}

\section{Multimodal Reasoning through Instruction Tuning}
\label{sec:mit}
Pre-trained LLMs are capable of complex reasoning in a few-shot manner when prompted with task-related exemplars or rationales \cite{brown2020language, wei2022chain, kojima2022large}.
One of the methods to achieve such ability of in-context learning (ICL) is through instruction tuning. 
With appropriate tuning,  the model is expected to enhance its ICL capabilities, \textit{e.g.} when prompted with examples of input-output pairs. 
However, there are several main challenges when the prompting modalities go beyond language. 
In this section, we first review current works on instruction tuning that facilitate ICL, and then the primary challenges and recent solutions in multimodal prompting.

\subsection{Definition of Instruction Tuning}
In recent years, instruction tuning has attracted significant attention due to the remarkable achievements of Large Language Models such as GPT-3\cite{brown2020language}, LLAMA~\cite{touvron2023llama}, Claude~\cite{bai2022training}, PaLM~\cite{chowdhery2022palm}. 
These models have demonstrated exceptional capabilities in various human-level understanding and reasoning tasks, including mathematics~\cite{hendrycks2021measuring,cobbe2021training}, coding~\cite{chen2021evaluating}, and most notably in generalizing to new tasks with zero or few-shot examples through instruction tuning.

Pretrained LLMs often struggle with generalizing to new tasks and aligning with users' intents, leading to untruthful, unhelpful, and unsafe responses. 
To mitigate these limitations, instruction tuning~\cite{wei2021finetuned,ouyang2022training} has been proposed.

Instruction tuning typically begins with a carefully collected dataset comprising thousands of  $\langle$\textbf{\textit{instruction}}, \textbf{\textit{response}}$\rangle$ pairs. 
In this process, LLMs are fine-tuned using this dataset to better align human intent with model behavior. 
It is worth noting that the tuning of model parameters is supervised by the loss function $\mathcal{L}_{\mathrm{it}}$, which is computed based solely on the \textbf{\textit{response}} tokens.
The loss function is defined as:
\begin{equation}
    \mathcal{L}_{\mathrm{it}} = -\sum_{x_i \in {\mathbf{response}}} \log p(x_i| x_{0:i-1}, \mathbf{instruction}).
\end{equation}

The pursuit of high-quality and diverse instruction-tuning datasets has been a key goal. 
A well-crafted \textbf{\textit{instruction}} usually includes clearly described question that represents human intent, along with any necessary background question context. 
In addition, the \textbf{\textit{response}} must be accurate and detailed. 
In the context of multi-modal instruction tuning, each example in the dataset consists of $\langle$\textbf{\textit{X-modality}}, \textbf{\textit{instruction}}, \textbf{\textit{response}}$\rangle$, where \textbf{\textit{X-modality}} typically represents images, videos, or audio. 
\tablename~\ref{tab:multimodal_tab} shows current literature and their training dataset, with some of these involving instruction tuning stage.
\begin{table}[htbp]
    \tiny
    \caption{Recent multimodal large language models, }
    \centering
    \begin{tabular}{l|p{1.3cm}|p{3.1cm}|p{3.0cm}|p{3.cm}}
    \toprule
    Methods & Visual Encoder &  Pre-training Stage & Multi-task learning Stage  &  Instruction Tuning \\
    \midrule
      OpenFlamingos &  CLIP-ViT-L/14~\cite{radford2021learning} & LAION-2B~\cite{schuhmann2022laion}, Multimodal C4~\cite{zhu2023multimodal}, Synthetic data  & - & - \\
    \midrule
    LLaMA-Adapter-v2 &  CLIP Vit-L/14 & GPT-4-LLM~\cite{peng2023instruction}, COCO    &-  & -  \\
    \midrule
      BLIP2 & CLIP ViT-L/14   &  COCO \cite{lin2015microsoft}, CC3M \cite{sharma2018conceptual}, CC12M \cite{changpinyo2021conceptual}, LAION400M \cite{schuhmann2021laion400m}, Visual Genome \cite{krishna2017visual},  &  -&-  \\
    \midrule
     InstructBLIP  & ViT-G/14  &  COCO, Visual Genome CC3M~\cite{sharma2018conceptual}, CC12M~\cite{changpinyo2021conceptual}, LAION400M~\cite{schuhmann2021laion400m}   & - & 26 publicly available vision-language datasets, transformed into the instruction tuning format\\
    \midrule
     LLaVA-1.5 & CLIP ViT-L/14 & LLaVA~\cite{liu2023llava}   & - & LLaVA665k (VQAv2~\cite{goyal2017making}, GQA~\cite{hudson2019gqa}, OK-VQA~\cite{marino2019ok}, AOK-VQA~\cite{schwenk2022okvqa}, OCRVQA~\cite{mishra2019ocr}, TextCaps~\cite{sidorov2020textcaps}, LLaVA150k~\cite{liu2023llava}, ShareGPT~\cite{shareGPT})\\
    \midrule
     mPLUG-owl2  & CLIP Vit-L/14  &LAION-400M, COYO~\cite{kakaobrain2022coyo-700m}, COCO, Laion-en,  DataComp~\cite{gadre2023datacomp} & - &VQAv2, OKVQA, OCR-VQA, GQA, A-OKVQA, RefCOCO~\cite{kazemzadeh2014referitgame}, Visual Genome, LLaVA150K, ShareGPT, SlimOrca~\cite{SlimOrca}  \\
        \midrule
      MiniGPT-v2  & EVA-G & GRIT-20M~\cite{peng2023kosmos2}, LAION, CC3M, SBU~\cite{ordonez2011im2text}, COCO caption, Text Captions~\cite{sidorov2020textcaps}, RefCOCO~\cite{kazemzadeh2014referitgame}, RefCOCO+, RefCOCOg~\cite{mao2016generation}, Visual Genome, GQA, VQAv2, OCR-VQA, OK-VQA, AOK-VQA &  GQA, VQAv2, OCR-VQA, OK-VQA, AOK-VQA, RefCOCO, RefCOCO+, RefCOCOg, Visual Genome, COCO caption, Text Captions & COCO Caption, Text Captions RefCOCO, RefCOCO+, RefCOCOg, Visual Genome, GQA, VQAv2, OCR-VQA, OK-VQA, AOK-VQA, LLaVA150k, Flickr30k~\cite{plummer2015flickr30k}, Multi-task conversation,Unnatural Instructions \\
    \midrule
    CogVLM & EVA2-CLIP-E & LAION-2B, 
COYO   & REC & LLaVA150k, LLaVAR~\cite{zhang2023llavar}, LRV-Instruction~\cite{liu2023mitigating}, in-house data \\
    \midrule
    QWen-VL-Chat & CLIP ViT-bigG & LAION-en, LAION COCO, DataComp, Coyo, CC12M, CC3M, SBU, COCO Caption, LAION-zh, In-house Data & LAION, DataComp, Coyo, CC12M, CC3M, SBU,
COCO, GQA, VGQA, VQAv2, DVQA~\cite{kafle2018dvqa}, OCR-VQA, DocVQA~\cite{mathew2021docvqa}, TextVQA, ChartQA~\cite{masry2022chartqa}, AI2D~\cite{hiippala2021ai2d}, GRIT, Visual Genome, RefCOCO, RefCOCO+, RefCOCOg, SynthDoG, Common Crawl, In-house Data & In-house data \\
     \bottomrule
    \end{tabular}
    \label{tab:multimodal_tab}
\end{table}

\subsection{Instruction Tuning for In-Context Learning}
In-context learning on Large Language Models has gained significant attention as a crucial indicator of models' reasoning capabilities. 
Models excelling in ICL should inherently possess strong analogical reasoning ability. 
LLMs with robust ICL capability can adapt to new tasks based on just a few examples, without the need for updating their weights. 
Recently, Multimodal Large Language Models built upon LLMs have also demonstrated such ICL capability. 
Frozen~\cite{tsimpoukelli2021multimodal} is the first MLLM to showcase ICL ability using a frozen LLM. 
Flamingo~\cite{alayrac2022flamingo} demonstrated strong ICL capabilities with a more powerful LLM and web-scale images interwoven with text for pretraining. 
OpenFlamingo~\cite{awadalla2023openflamingo} and IDEFICS~\cite{idefics} are open-source reproductions of Flamingo.
While they may not match Flamingo in every performance metric, they have demonstrated similar In-Context Learning (ICL) capabilities.

All ICL-capable MLLMs mentioned above are pretrained models.
During the instruction tuning stage, most MLLMs primarily focus on image-text pairs, which can gradually lead to a diminishing of ICL capabilities.
This is attributed to the challenges in constructing ICL-focused instruction tuning datasets. 
Recently, Otter~\cite{li2023otter} has developed the image-text-interleaved instruction tuning dataset MIMIC-IT~\cite{li2023mimicit}, which was then fine-tuned on OpenFlamingo. 
Another notable instruction fine-tuning dataset is MIC~\cite{zhao2023mmicl}, which includes in-context examples and is composed of 16 public datasets. 
In MIC, samples and in-context examples are constructed by randomly sampling from pre-defined templates.
While these advancements have equipped MLLMs with ICL abilities, challenges continue to exist in the area of multimodal prompting.

\subsection{Multimodal Prompting by Representation Learning}

Multimodal prompting, while opening up avenues to leverage rich information from non-linguistic modalities, often leads to catastrophic forgetting \cite{mccloskey1989catastrophic} due to disparities in information densities and a lack of high-quality paired data.
A key challenge in this domain is to derive LLM-compatible representations from non-linguistic inputs without altering the base LLM. 
This challenge has spurred research in multimodal representation learning, aiming to develop a unified prompt embedding space for various modalities. 
As an attempt to accommodate a secondary prompting modality and to harness the knowledge from the pre-trained LLM, Tsimpoukelli \textit{et al.} \cite{tsimpoukelli2021multimodal} train a vision encoder alongside a frozen LLM to transform vision prompts into an embedding sequence compatible with language embeddings. 
Additionally, Koh \textit{et al.} \cite{koh2023grounding} proposed a method for learning linear projections to convert embeddings between vision and language.
This approach enables their contextual-retrieval-based model FROMAGe to uniformly handle multimodal prompts.

\subsection{Multimodal Prompting by Exemplar Generation}

Prompting with in-context examples is a paradigm that unlocks the analogical reasoning ability of the pretrained LLMs for downstream tasks \cite{dong2022survey}. 
Although this approach is widely used in single-modality contexts~\cite{brown2020language, liu2021makes, lu2021fantastically, wu2022self}, there are relatively few studies on prompting with multimodal examples. 
A significant challenge in multimodal in-context learning is that multimodal LLMs often require contextual information to tackle new tasks, but acquiring in-context examples for every new task is not always feasible. 
Addressing this issue, Guo \textit{et al.} \cite{guo2023images} presented Img2LLM, a system capable of automatically generating LLM-agnostic exemplar prompts for visual question answering based on question images. 
This is achieved by first extracting answer candidates from generated captions and then formulating questions that correspond to these answers, thus creating a set of question-answer pairs suitable for new tasks.

\subsection{Multimodal Prompting by Model Interaction}
Combining the distinct capabilities of foundation models across different modalities for general multimodal tasks remains challenging, often requiring computationally expensive joint pretraining or fine-tuning. 
Zeng \textit{et al.} \cite{zeng2022socratic} introduced the concept of multimodal prompting.
This approach facilitates the exchange of information between different modalities through language prompts.
They demonstrate that by thoughtfully crafting interactive processes and prompt templates for specific multimodal tasks, these foundation models can perform new tasks in a zero-shot manner, eliminating the need for additional training. Wang \textit{et al.} \cite{wang2022language} similarly uses text as the interface to realize multimodal in-context learning without any training.
\section{Applications}
In this section, we mainly cover Embodied AI and Multimodal Agent to discuss the applications of MLLM models.

\subsection{Embodied AI}
\label{sec:emboded}

Embodied AI has become increasingly popular with the advancement of powerful AI models. 
The Embodied AI workshop at the conference on Computer Vision and Pattern Recognition (CVPR) offers this definition:
\begin{quote}
	\textit{
    Minds live in bodies, and bodies move through a changing world. The goal of embodied artificial intelligence is to create agents, such as robots, which learn to creatively solve challenging tasks requiring interaction with the environment. 
    \begin{center}
      \url{https://embodied-ai.org/}
    \end{center}}
\end{quote}
This field requires the integration of multiple technologies and knowledge areas, including Computer Vision, Speech, Natural Language Processing, and Control Systems.
These are essential for enabling an agent to see, talk, listen, and interact with its environment. 
More importantly, it also demands that the agent possess reasoning abilities, allowing it to plan actions based on environmental cues and adapt its behavior in response to feedback from its surroundings. 

\begin{figure}
    \centering
    \includegraphics[width=.9\textwidth]{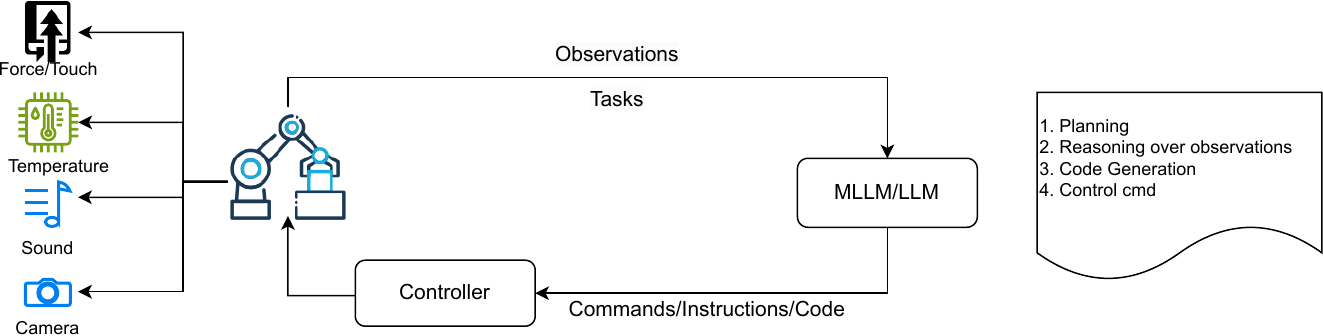}
    \caption{
    	 Illustration of the interaction between an Embodied Agent and MLLM/LLM models. Here, pretrained models generate commands and planning outcomes for the robot. In turn, the robot provides environmental observations to the MLLM models, aiding in refining their reasoning for future tasks.}
    \label{fig:embodied:abs}
\end{figure}

Robotics stands as the foremost representation of embodiment AI agents. 
The recent advancements in MLLMs shed light on robotic manipulation by leveraging the commonsense knowledge from the visual world and the reasoning ability of LLMs.
\figurename~\ref{fig:embodied:abs} illustrates the interaction loop between an Embodied Agent and an MLLM.
Specifically, this commonsense knowledge assists the Embodied agents in tasks like planning, complex reasoning, code generation, and the creation of control commands. 
These generated instructions, in various formats, are then provided to the control unit for the generation and execution of actions.
This section provides a summary of the latest applications of MLLMs in embodied AI, with a focus on their reasoning capabilities.

\subsubsection{Tasks and Challenges of Embodied AI Reasoning}
Research tasks in Embodied AI are broadly classified into three categories: visual exploration, visual navigation, and embodied question answering~\cite{duan2022survey, savva2019habitat}. 
These tasks require the embodied agent to perceive its environment, plan and execute control strategies, and interact with humans or the environment. 
A primary challenge in this field is adapting agents to out-of-domain environments, which demands robust commonsense knowledge and reasoning abilities from the agent. 
This aligns with the recent advancements in pretrained large models, including both LLMs and MLLMs.
The subsequent subsections will delve into how MLLMs can be applied to enhance the capabilities of Embodied Agents.

\subsubsection{Direct Reasoning with MLLM Knowledge}
Early methods of employing MLLMs primarily centered on the direct use of pre-trained LLM and MLLM models. 
These models typically incorporate commonsense knowledge acquired from extensive datasets.
Observations gathered by the Embodied Agent are provided to these models, leveraging their pretrained reasoning abilities for task planning and problem-solving.
The principal strategy involves decomposing the embodied tasks into smaller, more executable steps,  allowing the pre-trained model to efficiently tackle each intermediate stage. 
Initially, these studies focused less on additional training of the model and more on the task decomposition and the efficient coordination of subtasks. 

\paragraph{Reasoning with Separate Components.}
CLIPort~\cite{shridhar2022cliport} represents an early endeavor to harness the semantic knowledge of pretrained models trained on large-scale datasets. 
In particular, it utilizes CLIP~\cite{radford2021learning} to assist broad semantic understanding.
Featuring a two-stream architecture, their end-to-end framework is capable of solving various tasks without the need for explicit concept learning for each task.

Later, Socratic Models (SMs) are proposed in~\cite{zeng2022socratic}, where each pretrained model is assumed to contain commonsense knowledge from distinct domains. 
By combining these models, they can collectively address tasks across various domains. 
Specifically, different models can be composed by using language prompts as the bridge. 
The proposed Socratic Models employ a multimodal prompting strategy to combine different VLM, LM models. 

In a related study, LM-Nav~\cite{shah2023lm} leverages pretrained models for robotic navigation. 
The main idea is to employ language as the bridge to break down the instructions. 
The process involves employing GPT-3 for identifying textual landmarks for instructions, using CLIP for landmark grounding, and optimizing a probabilistic objective by inferring a joint likelihood over landmarks and nodes. 
Again, this method requires no further model fine-tuning.

More recently, Dasgupta \textit{et al.}~\cite{dasgupta2023collaborating} explored the Embodied AI agent concept within a \textbf{Planner-Actor-Reporter} paradigm. 
The ``Planner'' interprets human instructions and directly translates them into sequences of simple steps to be executed by the ``Actor'', which can be implemented by a pretrained LLM.
The ``Reporter'' observes the environment, gathers information, and reports it to the ``Planner''. 
The information will be appended to the previous instructions to form the new prompt.
Their experiments highlight the advantages of this paradigm. 
They also suggest that the ``Reporter'' can be trained within a Reinforcement Learning (RL) framework.

\paragraph{Reasoning with External World Model.}
Currently, LLM has a limit of window context. 
To execute a task that involves the states or facts from the distant past, this work proposes to use a world state to represent the knowledge~\cite{yoneda2023statler}.
In particular, the proposed approach attempts to use memory or a world model to record the key states of the current environment. 
The mechanism includes a world state reader and world state writer, which utilizes the states as prompts for LLM to solve a complex task and update the world state memory according to new observations. 
Experimental results highlight the effectiveness of this model, surpassing existing methods in certain aspects.
However, there are areas for improvement. The method still requires task-specific prompt definition, lacks visual feedback, and the world state may be updated inaccurately.

\paragraph{Reasoning with Feedback.}
Inner Monologue~\cite{huang2023inner} attempts to imitate human thought processes in \textit{natural languages}.
Inspired by human cognition, this work proposes a method for solving robotic tasks by using languages as a bridge. 
Robots are equipped with ``short-horizon skills'', enabling them to execute basic tasks or commands described in simple language.
Additionally, a Large Language Model is employed as a planner.
This LLM aims to fulfill high-level user commands by breaking them down into a sequence of skills that the robot can perform.

Furthermore, the system can also incorporate environmental ``textual'' feedback into its prompts, a process described as an  ``inner monologue''. 
This feedback includes success detection, object and scene description, Visual Question Answering (VQA), and human feedback. 
The authors highlight the emergent capabilities of their system. 
Most of the reasoning and re-planning are facilitated by the LLM. 
In particular, the authors attribute the success of this reasoning to the incorporation of environment feedback into the model's decision-making process.
\paragraph{Reasoning with Policy Generation.}
The research presented in~\cite{liang2023code} further studies the application of language model programming for robotic control.
Moving beyond the direct use of the pretrained LLM's language capabilities, this study also involves prompting an LLM with coding ability to generate policy code.

This approach further expands the scope of reasoning tasks to include arithmetic and novel instructions. 
Specifically, this study uses OpenAI Codex code-davinci-002. 
It uses prompts, supplemented with examples, to guide the LLM in generating additional functions that adhere to user instructions. 

Subsequently, the generated code is executed using the robot's control primitives. 
As discussed in the paper, this system effectively utilizes LLMs and code to enhance its reasoning capacity. 
However, it is constrained by the limitations of the available API and control primitives. 
Additionally, the system lacks a mechanism to assess the feasibility of a given instruction.
\subsubsection{Improved Reasoning with Robotic data}
Collecting robot trajectories with crowd-sourced annotations is challenging, often resulting in datasets of limited scale. 
The authors from~\cite{xiao2022robotic} propose data driven instruction augmentation for language-conditioned control (\textbf{DIAL}). 
Their approach consists of several key steps: 1) Fine-tuning a Vision-Language Model (VLM) on a small,  manually annotated dataset of robot manipulation trajectories; 2) Gathering a larger, unannotated dataset;  
3) Using the fine-tuned VLM to annotate this newly collected dataset; 
4) Retraining the VLM on both the original and the newly labeled dataset. 

In the re-labeling step, the authors utilize GPT-3 to expand the range of human instruction candidates.
They explore various strategies for selecting appropriate instructions for given robot trajectories. 
Overall, DIAL demonstrates more capabilities in solving novel tasks. 
It can also be used to distill Internet-scale vision language models into embodied agents~\cite{sumers2023distilling}.

Another approach involves integrating discrete text tokens directly into the training set as natural language tokens~\cite{brohan2023rt}, further enriching the model's training data.
\subsubsection{Embodied Foundation Model}
\paragraph{PaLM-E~\cite{driess2023palme}: the Largest Reported Embodied Multimodal Large Language Model.}
The primary goal is to transform inputs from various modalities into formats compatible with LLMs, thereby enabling embodied agents to interact with environments across different modalities. 
The paper evaluates various methods for converting these modalities into LLM inputs. 
In particular, the object scene representations~\cite{sajjadi2022object} demonstrate its effectiveness even with limited training data in robotic applications. 
The largest PaLM-E-562B demonstrates emergent abilities, including zero-shot multimodal chain-of-thought reasoning. 
This is largely due to its ability to transfer knowledge from general visual-language understanding to tasks related to embodied AI.

However, the reasoning or planning skills of PaLM-E are integrated with separate low-level policies that generate robot control commands for execution. 
In this work, the authors employ RT-1~\cite{brohan2022rt} for these low-level policies.
RT-1 processes visual observations and language instructions, converting them into robot control commands. 
\paragraph{RT2: Vision-Language-Action (VLA) Model.}
Subsequently, the following approach, RT-2~\cite{brohan2023rt}, directly utilizes a pretrained VLM model to generate robot control commands. 
To achieve this, they apply the action discretization pipeline. 
This pipeline transforms each dimension of the seven-dimensional continuous action space into 256 uniformly distributed discrete actions. 

Following this transformation, the RT-2 model undergoes fine-tuning with both robotics data and original web data. 
This approach shows better generalization than models fine-tuned solely with robotic-only data.

For the base VLM, the authors employ two of the largest VLM models available: PaLI-X~\cite{chen2023pali} and PaLM-E~\cite{sajjadi2022object}. 
Their results prove that after fine-tuning the VLA models with robot trajectory data, RT-2 demonstrates several emergent abilities.
These include enhanced reasoning capabilities with novel objects and improved interpretation of human instructions. 
More interestingly, after only a few hundred gradient steps of fine-tuning, RT-2 acquires chain-of-thought reasoning abilities.
The data augmentation process introduces an additional ``Plan'' step, wherein the model is required to plan its actions before executing them. 
This seemingly simple addition significantly enhances the model's ability to comprehend and respond to more complex commands. 
\subsection{MLLMs for Tool Usage}
\label{sec:tool}
\subsubsection{Tool Usage in LLM}
Recent studies have explored the potential of using external tools to augment LLMs' capabilities for complex tasks \cite{yao2023react}. 
Research \cite{parisi2022talm,schick2023toolformer,paranjape2023art} has demonstrated that
supplementing LLMs with tools, such as calculators, search engines, translation systems, calendars,
or even API calls to other models can help solve tasks beyond their inherent capabilities.

The integration of tools into LLMs generally involves two primary phases: tool selection or creation, and processing the results obtained from these tools~\cite{qin2023tool,mialon2023augmented,yang2023foundation}. 
Tool selection, a pivotal phase within tool learning, can involve allowing the LLM itself to make decisions based on input queries \cite{qin2023toolllm,qian2023creator,nakano2022webgpt,yao2023react,driess2023palme}.
Alternatively, it might include the use of predefined tools specifically designed for particular scenarios\cite{schick2023toolformer,lu2023chameleon,hsieh2023tool,thoppilan2022lamda}, such as retrieval systems. 

When the model is required to autonomously determine which tools to employ, it necessitates the model to understand the characteristics of these tools and make accurate selections via their reasoning abilities. 
The majority of approaches tend to decompose a complex task into a series of subtasks interacting with tools in sequential or tree structures, such as Chain of Thoughts (CoT)~\cite{wei2022chain,kojima2022large}, Tree of Thoughts (ToT)~\cite{yao2023tree} and Program of Thoughts (PoT)~\cite{chen2022program}. 

A simple way to teach language models to use tools is by instruction~\cite{yao2023react, hsieh2023tool, paranjape2023art}. 
The description of tools is able to teach language models to use tools~\cite{hsieh2023tool} in a zero-shot fashion. 
Language models can also learn to use tools from few-shot examples. 
This in-context tool learning can range from retrieval tasks to enhancing the model's ability to generalize to new challenges. 
For example, Art~\cite{paranjape2023art} demonstrates this by selecting examples of multistep reasoning and tool usage from a task library. 
This approach significantly improves upon traditional few-shot prompting and automated Chain of Thought (CoT) methods, particularly in unseen tasks within the BigBench and MMLU benchmarks. 

Some studies have shown that LLMs can learn to utilize tools by fine-tuning on the tool-specific corpora \cite{parisi2022talm,schick2023toolformer,patil2023gorilla}. 
TALM \cite{parisi2022talm} and Toolformer \cite{schick2023toolformer} both adopt a self-supervised method, starting with a few tool demonstrations and then enabling the language model to autonomously generate additional data involving tool usage. 
TALM is designed to use APIs for search and calculation, applying these functions to enhance language models' capabilities in question-answering tasks. 
Toolformer, on the other hand, utilizes a broader range of tools, including a calculator, a QA
system, a search engine, a translation system,
and a calendar, leading to improved zero-shot performance across various downstream tasks.

Gorilla \cite{patil2023gorilla} introduces APIBench, a comprehensive dataset comprising APIs from HuggingFace, TorchHub, and TensorHub.
The study demonstrates that a fine-tuned LLaMA-based model, trained with this dataset, significantly surpasses GPT-4 in generating API calls, showcasing its enhanced performance.

Recent research~\cite{qian2023creator,chen2022program} has expanded the capabilities of LLMs beyond just utilizing existing tools; it now includes enabling LLMs to create tools through programming. 
This method uses programming languages to tackle mathematics tasks, thereby enhancing the models' capacity for accurate mathematical computation and improving logical task planning.

\subsubsection{MLLMs as Tool}

\begin{table}[!ht]
    \centering
    \caption{Comparison of Works Using MLLM as Tools. For Modalities Integrating Works, we list the tool domain instead of enumerating all individual tools.}
    \resizebox{\linewidth}{!}{
    \setlength{\tabcolsep}{1.0 mm}{
    \begin{tabular}{l|c|c|c}
    	\toprule
    	 \rowcolor{gray!20}    & & & \\
    	\rowcolor{gray!20}\multirow{-2}{*}{Application} & \multirow{-2}{*}{Work} & \multirow{-2}{*}{ Reasoning method } & \multirow{-2}{*}{Tools}\\ 
        \midrule
        Compositional Visual Reasoning & VISPROG \cite{gupta2023visual} & Program & OWL-ViT, DSFD, MaskFormer, CLIP, ViLT, Stable Diffusion\\ \hline
        \multirow{2}{*}{Multimodal Dialogue} & Visual ChatGPT \cite{wu2023visual} &  CoT &  BLIP, Stable Diffusion, Pix2Pix, ControlNet\\ 
        & MM-ReAct \cite{yang2023mm} &  ReAct & image captioning, tagging, celebrity recognition, OCR, Web \\ \hline 
        \multirow{2}{*}{Modalities Integrating} & Chameleon \cite{lu2023chameleon} &  Program & Modalities: image, web, knowledge, math, table \\ 
        & TaskMatrix.AI \cite{liang2023taskmatrix}  & CoT & Modalities: image, office, cloud, robot \\ \hline
    \end{tabular}
    \label{MLLMs as Tool}
    }
    }
\end{table}

MLLMs can function as tools to empower language models in handling various modalities and addressing challenging visual reasoning tasks. We briefly review these works from different application perspectives, as illustrated in \tablename~\ref{MLLMs as Tool}.

\paragraph{Compositional Visual Reasoning.} 
While current MLLM models demonstrate proficiency in visual tasks such as image classification and captioning, they encounter difficulties with more complex visual operations, like visual editing, due to inherent requirements for reasoning and planning. 
For instance, replacing a person's face with a cat's face in an image involves face detection, person recognition, cat face generation, and merging the new face into the original image. 
Critical reasoning is required to break down the tasks and develop a detailed execution plan.
MLLMs can play a pivotal role in carrying out these decomposed steps.

To solve compositional visual reasoning tasks, VISPROG \cite{gupta2023visual} utilizes MLLMs as a toolset within a modular and interpretable neuro-symbolic system.
This system, designed for comprehensive visual understanding and manipulation, leverages GPT-3 to generate programs through in-context learning.
These programs coordinate different modules, such as OWL-ViT \cite{minderer2205simple} for open vocabulary localization, DSFD \cite{li2019dsfd} for face detection, MaskFormer \cite{cheng2021per} for segmentation, CLIP \cite{radford2021learning} for image selection and classification, ViLT \cite{kim2021vilt} for visual question answering, and Stable Diffusion \cite{rombach2022high} for object replacement. 
Additional tools such as Python Imaging Library\footnote{\url{https://pypi.org/project/pillow/}} (Pillow) and OpenCV\footnote{\url{https://opencv.org/}} are also used for image manipulation. 
VISPROG can handle various complex visual reasoning tasks, including compositional VQA, zero-shot reasoning on image pairs, factual knowledge object tagging, and image editing using natural language.

 \paragraph{Multimodal Dialogue Reasoning.} 
 Following dialog instructions can become more complex than single-step instructions due to their context-dependent nature.
 Multi-turn dialogue understanding requires reasoning based on dialog history. 
 MLLMs can serve as tools for LLMs to facilitate dialogs that incorporate visual inputs and outputs, drawing on the reasoning capabilities of LLMs to interpret dialog instructions.

Visual ChatGPT~\cite{wu2023visual}, incorporating various MLLM, enables users to interact with ChatGPT using both text and images.
This system harnesses MLLMs such as BLIP \cite{li2023blip2}, Stable Diffusion \cite{rombach2022high}, and other visual foundation models like Pix2Pix \cite{rombach2022high}, ControlNet \cite{zhang2023adding}.
It is capable of performing tasks like image captioning, image question answering, and image editing during conversations. 
Chain-of-Thought (CoT) \cite{wei2022chain} reasoning method is used for case that requires collaboration between multiple visual models. 

Additionally, MM-ReAct \cite{yang2023mm} utilizes the ReAct \cite{yao2023react} planning method alongside a suite of image and video tools for a stronger capacity of visual tasks. 
These tools cover a wide range of functions, including image captioning, image tagging, dense captioning, celebrity recognition, receipt scanning, OCR, and Bing search.
This enables MM-ReAct to handle difficult tasks such as Visual Math and Text Reasoning, Visual-Conditioned Jokes/Memes, and Visual Planning.

\paragraph{Integrating Modalities Beyond the Visual Domain.}

MLLMs are usually customized for specific modalities, such as images, which poses a challenge when trying to integrate their usage across varied domains like mathematics, the web, tables, and more. 
For instance, solving a math problem depicted in an image requires the visual model to first interpret the image and then invoke the math model for a solution. 
Addressing this challenge entails fostering reasoning and planning among various modality models. 
By conceptualizing all the modality models as tools and leveraging the reasoning capabilities of LLM, recent efforts aim to integrate additional modalities beyond the visual domain, such as web content, tables, office documents, and robotic data. 
This approach contributes to the development of a more powerful multimodal model.

Chameleon \cite{lu2023chameleon} exemplifies this approach.
It can utilize a variety of tools, such as existing vision models, web search engines, Python functions, and heuristic-based modules.
Capable of handling tasks across image, web, knowledge, math, and table domains, Chameleon synthesizes programs by combining these tools to coordinate their sequence through an LLM-driven planner for multistep tasks. 

TaskMatrix.AI \cite{liang2023taskmatrix} provides a comprehensive set of APIs catering to a wide range of tasks.
These include image editing, image-based question answering, image captioning, text-to-image conversion, and transformations like image-to-sketch/depth/hed/line. 
Furthermore, TaskMatrix.AI extends its utility to conversational robotics and IoT device control, offering APIs for tasks like robotics pick, move, and put, car air conditioner control, TV operation, and music playback. TaskMatrix.AI utilizes the CoT \cite{wei2022chain} reasoning method to device tool usage plans.
\section{Multimodal Reasoning Benchmark Result Analysis}
\label{sec:eval}

\begin{table*}[htbp]
    \centering
    \caption{Evaluation results for various MLLM on MMVet and MMMU (validation set). Open-source models best performances are indicated with underlines.}
    \setlength{\tabcolsep}{6pt}
    \begin{tabular}{l|c|c}
    \toprule
     \rowcolor{gray!20}    & & \\
     \rowcolor{gray!20}\multirow{-2}{*}{MLLMs} & \multirow{-2}{*}{MMVet} & \multirow{-2}{*}{MMMU} \\
    \midrule 
         
         MiniGPT-4 \cite{zhu2023minigpt}   & 24.4 & 26.8\\
         OpenFlamingo-v2~\cite{awadalla2023openflamingo}   & 24.8 & 28.7\\
         LLaMA-Adapter V2~\cite{gao2023llamaadapter}& 31.4 & 29.8\\
         CogVLM-Chat~\cite{wang2023cogvlm} & \textbf{\underline{52.8}} & 32.1 \\
         Otter~\cite{li2023otter}  & 24.6 & 32.2\\
         mPLUG-Owl2~\cite{ye2023mplugowl2}  & 36.3 & 32.7\\
         BLIP-2~\cite{li2023blip2}   & 22.4 & 35.4\\
         InstructBLIP~\cite{dai2023instructblip}   & 26.2 &35.7\\
        QWen-VL-Chat~\cite{bai2023qwenvl} & - & 35.9\\
        LLaVA-1.5~\cite{liu2023llava} & 36.3 & 36.4\\
        InfiMM-LLaMA-13B~\cite{infimm-v1} & 39.2 & \textbf{\underline{39.1}} \\
        \midrule
        GPT-4V~\cite{openai2023gpt4}  & \textbf{67.7} & \textbf{55.7} \\
        
    \bottomrule
    \end{tabular}
    \label{tab:main_result2}
\end{table*}

In this section, we briefly summarize the performance of existing Multimodal Large Language Models on the evaluation benchmarks outlined in Sec.~\ref{sec:mm_bench}.
Our focus is primarily on accessing the reasoning capabilities of these models.

\tablename~\ref{tab:main_result2} presents a summary of the results from both MM-Vet and MMMU datasets. 
MM-Vet~\cite{yu2023mmvet} is an open-ended QA benchmark designed for evaluating the comprehensive capabilities of MLLMs. 
According to their leaderboard, GPT-4V has achieved the best score among all MLLMs. 
The GPT-4 driven MM-ReAct~\cite{yang2023mmreact} also performs exceptionally well, particularly in reasoning tasks like mathematics and spatial reasoning.
This indicates that MLLMs' reasoning capabilities are heavily influenced by their underlying LLM.
Among the open-source models, CogVLM-Chat has achieved the highest score. Compared with other public models, the response of IFT dataset used in CogVLM-Chat are mostly long format and go through further manual correction. Such data aligns better with open-ended evaluation, thus achieved highest score on MM-Vet.

MMMU~\cite{yue2023mmmu} is a benchmark designed to evaluate MLLMs on college-level subject questions. 
The leaderboard reveals intriguing results: proprietary MLLMs, like Gemini Ultra/Pro~\cite{team2023gemini}, GPT-4V~\cite{openai2023gpt4v}, and Qwen-VL-PLUS~\cite{qwenvlplus2023} achieved much better scores than most open-source MLLMs. 
Notably, the scores of these open-source MLLMs are even lower than those of LLMs aided by OCR tools and image captions generated by LLaVA. 
MLLMs that reserve LLMs' capability, like Blip2/Instruct-Blip and InfiMM-LLaMA-13B achieved good performance, suggesting that MMMU requires model to retain knowledge obtained during LLM training and most other open-source MLLMs may experience catastrophic forgetting of language capability after visual IFT.

\begin{table*}[t]
	\centering
	\caption{Evaluation results for various MLLMs on InfiMM-Eval. The top performances achieved by open-source models are indicated with underlines for easy reference.}
	\resizebox{\textwidth}{!}{
		\setlength{\tabcolsep}{6pt}
		\begin{tabular}{l|ccc|cc|c}
			\toprule
			
			\rowcolor{gray!20}   & \multicolumn{3}{c|}{Reasoning Category} &\multicolumn{2}{c|}{Reasoning Complexity}&  \\
			\rowcolor{gray!20}\multirow{-2}{*}{MLLMs} & Deductive & Abductive & Analogical & Moderate & High & \multirow{-2}{*}{Overall} \\
			\midrule 
			OpenFlamingo-v2~\cite{awadalla2023openflamingo}   & 8.88 & 5.3 & 1.11 & 9.47 & 4.72& 6.82 \\
			MiniGPT-v2~\cite{zhu2023minigpt}   & 11.02 & 13.28 & 5.69 & 14.45 & 7.27& 10.43 \\
			Fuyu-8B~\cite{fuyu-8b}  & 16.42 & 21.49 & 7.78 & 23.06 & 9.91& 15.7 \\
			BLIP-2~\cite{li2023blip2}   & 22.76 & 18.96 & 7.5 & 24.05 & 14.18& 19.31 \\
			InternLM-XComposer-VL~\cite{zhang2023internlmxcomposer}  & 26.77 & 35.97 & 18.61 & 39.13 & 17.18& 26.84 \\
			InstructBLIP~\cite{chung2022scaling} & 27.56 & 37.76 & 20.56 & 40.64 & 18.09& 28.02 \\
			LLaMA-Adapter V2~\cite{gao2023llamaadapter} & 28.7 & 46.12 & 22.08 & 41.33 & 21.91& 30.46 \\
			Otter~\cite{li2023otter}   & 22.49 & 33.64 & 13.33 & 35.79 & 12.31 & 22.69 \\
			mPLUG-Owl2~\cite{ye2023mplugowl2}  & 23.43 & 20.6 & 7.64 & 28.79 & 13.18 & 20.05 \\
			IDEFICS-9B-instruct~\cite{laurenccon2023obelisc}  & 22.99 & 34.63 & 20.56 & 34.45 & 16.73& 24.53 \\
			Emu~\cite{sun2023generative}  & 28.9 & 36.57 & 18.19 & 36.18 & 22.0 & 28.24 \\
			LLaVA-1.5~\cite{liu2023llava} & 30.94 & 47.91 & 24.31 & 47.4 & 21.0& 32.62 \\
			CogVLM-Chat~\cite{wang2023cogvlm} & 36.75 & 47.88 & 28.75 & 55.67 & 22.5& 37.16 \\
			Qwen-VL-Chat~\cite{bai2023qwenvl}  & 37.55 & 44.39 & \textbf{\underline{30.42}} & 46.61 & \textbf{\underline{30.09}} & 37.39 \\
			SPHINX-v2~\cite{lin2023sphinx}  & \textbf{\underline{42.17}} & 49.85 & 20.69 & 54.85 & 27.31 & 39.48 \\
   InfiMM-LLaMA-13B~\cite{infimm-v1}  &39.9 & \textbf{\underline{53.18}} & \textbf{\underline{31.94}} & \textbf{\underline{55.85}} & 28.7 & \textbf{\underline{40.7}} \\
			\midrule
			GPT-4V~\cite{openai2023gpt4}  & \textbf{74.86} & \textbf{77.88} & \textbf{69.86} & \textbf{93.98} & \textbf{58.98} & \textbf{74.44} \\
			\midrule 
			\bottomrule
	\end{tabular}}
	\label{tab:main_result}
\end{table*}

InfiMM-Eval\cite{han2023infimmeval} is an open-ended question-answering benchmark designed to evaluate MLLMs' visual reasoning capability. 
In contrast to MMMU, which heavily relies on the knowledge capacity of the underlining LLM, InfiMM-Eval focuses more on the fundamental unified logic capabilities of MLLMs. 
The evaluation results on InfiMM-Eval are shown in \tablename~\ref{tab:main_result}.
Similar to other benchmarks, GPT-4V has shown a significant lead over other MLLMs. 
Among the open-source models, SPHINXv2~\cite{lin2023sphinx}, Qwen-VL-Chat~\cite{bai2023qwenvl}, and CogVLM-Chat~\cite{wang2023cogvlm} rank as the top three.
While Qwen-VL-Chat and CogVLM-Chat have achieved similar overall scores, their performance varies across different levels of reasoning complexity. 
Since CogVLM-Chat employs similar underline LLMs with other models, it is hypothesized that Qwen-VL-Chat's superior performance in highly complex questions may stem from the Qwen language model, whereas CogVLM-Chat's strengths in moderately complex questions are likely due to its enhanced detailed perception capability. 

In reviewing the top-performing open-source models on reasoning-focused benchmarks such as SPHINX-v2, QWen-VL-Chat,  CogVLM-Chat and InfiMM-LLaMA-13B, we summarize the following common recipes to facilitate future research:
\begin{itemize}
    \item Unfreezing the language model at a certain training stage is instrumental in enhancing multimodal reasoning abilities. 
    For example, SPHINX-v2 and QWen-VL-Chat unfreeze their language models during the pretraining and instruction tuning stages, respectively. 
    CogVLM, on the other hand, trains visual experts to modify the language model's response to visual input and also unfreeze the language model during fine-tuning. 
    While tuning the language model is expected to improve multimodal performance, careful control of training parameters such as the number of training steps and the learning rate is crucial to prevent catastrophic forgetting.
    \item Improving visual representations is key to better multimodal performance. 
    This can be achieved by increasing the resolution of images for finer details, as done in QWen-VL-Chat and CogVLM, or by employing stronger or multiple visual encoders, like in SPHINX-v2. 
    Another effective method is fine-tuning the visual encoder during the MLLM training process, as demonstrated by QWen-VL-Chat.
    \item The multi-task supervised learning stage significantly contributes to the superior performance of top models like QWen-VL-Chat, CogVLM, and InfiMM-LLaMA-13B. Each of these models employs a three-stage training process, demonstrating the value of this approach. Notably, InfiMM-LLaMA-13B, while using the same instruction dataset as LLaVA-1.5, showcases improved reasoning abilities across various benchmarks. The multi-task supervised learning stage can be achieved by utilizing datasets of higher quality than the pretraining stage, yet containing less conversational data compared to the instruction tuning stage. Additionally, CogVLM achieves impressive performance on the mm-vet, utilizing unknown in-house data during the instruction tuning stage.
\end{itemize}

Besides these general strategies, we also observed dataset-specific trends and variances:
\begin{itemize}
    \item As shown in \tablename~\ref{tab:infimm_ab}, instruction tuning significantly enhances performance on InfiMM-Eval, arguably the most representative dataset for evaluating multimodal reasoning. 
    The importance of instruction tuning for MLLMs is also verified by the fact that MLLMs usually take instruction tuning as their final training stage. 
    \item There are significant differences in model rankings between InfiMM-Eval and MMMU. 
    For example, BLIP-2, which performs relatively poorly on InfiMM-Eval, excels on MMMU, surpassing models such as CogVLM. 
    Due to a range of factors, including the types of evaluation datasets, language model architectures, and training methodologies, it is challenging to pinpoint the exact cause of these discrepancies. 
    However, it is important to highlight the potential for performance gaps across various evaluation datasets.
\end{itemize}

Note that our analysis is more of a case-study approach, focusing on the findings from top-performed models, especially in the context of the reasoning-focused InfiMM-Eval dataset.
Therefore, these insights may not be universally applicable across various training data source, language models, and their respective scales.

\begin{table*}[t]
    \centering
    \caption{Evaluation results for MLLMs before and after instruction tuning on InfiMM-Eval.}
    \setlength{\tabcolsep}{6pt}
    \begin{tabular}{l|ccc|cc|c}
    \toprule

     \rowcolor{gray!20}   & \multicolumn{3}{c|}{Reasoning Category} &\multicolumn{2}{c|}{Reasoning Complexity}&  \\
     \rowcolor{gray!20}\multirow{-2}{*}{MLLMs} & Deductive & Abductive & Analogical & Moderate & High & \multirow{-2}{*}{Overall} \\
    \midrule 
         Qwen-VL-7B~\cite{bai2023qwenvl}  & 24.06 & 24.39 & 9.03 & 32.33 & 12.69 & 21.32 \\
         Qwen-VL-7B-Chat~\cite{bai2023qwenvl}  & 32.25 & 48.18 & 24.03 & 46.90 & 22.78 & 33.44 \\

    \midrule 
    \bottomrule
    \end{tabular}
    \label{tab:infimm_ab}
\end{table*}

\section{Conclusions and Future Directions} 
\label{sec:conclusion}
The concept of reasoning ability is pivotal in the quest for Strong Artificial Intelligence (Strong AI) or Artificial General Intelligence (AGI), a goal that has been pursued across various scientific disciplines for several decades.
With recent advancements in both models and evaluation benchmarks, there is a growing discourse on the reasoning abilities exhibited in current LLMs and MLLMs.
In this work, we present the different types of reasoning and discuss the models, data, and evaluation methods used to measure and understand the reasoning abilities demonstrated in existing studies.
Our survey aims to provide a better understanding of our current standing in this research direction and hopes to inspire further exploration into the reasoning abilities of future work.

In addition, we wish to outline some potential research directions to improve the reasoning capabilities of current MLLMs. 
The following discussions include our summaries regarding pretraining, alignment and evaluation benchmarks:
\paragraph{MLLM Architectures.}
Current MLLM architectures still have the following fundamental limitations in achieving human-level multimodal reasoning ability. 
The first limitation comes from the image resolution. 
Current Multimodal Large Language Models typically operate on images with a fixed, preset resolution. 
However, in real-world scenarios, images can come in a variety of resolutions.
The hallucination problem of MLLM also possibly stems from the MLLM architecture design. 
With additional modalities as input, it is more complicated to diagnose the source of the hallucinations in MLLMs.
It becomes challenging to determine whether these hallucinations arise from an incomplete perception of visual signals or from biases and spurious correlations ingrained in language models. 
Therefore, it is important to design MLLM architectures that can inherently minimize or avoid hallucinations caused by inaccurate perception.

\paragraph{Efficiency and Scalability of MLLM Training Recipes.}
As shown in \figurename~\ref{fig:mllm:MLLM_example}, different MLLMs utilize vastly varying amounts of data in each training stage.
However, performance comparisons do not show proportional differences (\tablename~\ref{tab:main_result} and \tablename~\ref{tab:main_result2}).
Understanding the relationship between data volume and model scale at each training stage is crucial.
According to recent studies~\cite{lin2021vx2text,liu2023improved,lin2023towards}, the pretraining stage for aligning vision-text representation may not require extensive computational resources.

\paragraph{Long-context Support.}
Existing MLLMs often serve in short-context situations and lack the ability to handle diverse real-world, long-context scenarios. 
These scenarios range from comprehending lengthy documents and papers to more complex tasks like understanding videos or movies. 
While most MLLMs employ attention mechanisms, the challenge with long-context tasks is the need to minimize computational complexity and maintain focus as the prompt length increases. 
A significant gap exists in research on multimodal long-context tasks, largely attributable to the absence of appropriate datasets and evaluation benchmarks that are equipped to tackle these specific challenges.

\paragraph{Instruction Fine-tuning (IFT) Data.}
An enhanced approach to instruction fine-tuning data is necessary, given that most of the existing public MLLM instruction datasets are insufficient in developing reasoning abilities for multimodal tasks.
Currently, there is a lack of ablation studies exploring the type of data that could boost multimodal reasoning capabilities.
Therefore, the creation of future instruction datasets demands careful consideration to more effectively improve reasoning capabilities in multimodal contexts.

\paragraph{Reinforcement Learning for Multimodal Reasoning.}
Inspired by the success of InstructGPT~\cite{ouyang2022training}, leveraging reinforcement learning algorithms~\cite{schulman2017proximal} to align model predictions and human preference has been extensively studied~\cite{bai2022training, rafailov2023direct}. 
However, reinforcement learning with human feedback (RLHF) has been just applied to MLLM~\cite{sun2023aligning} and there are many exciting open research problems. 
For example, given the complexity of multiple modalities, it is important to collect preference data effectively and scalably. 
With sufficient data, it is also unclear whether existing learning algorithms can handle the multimodal tasks or not. 
Another example is how to develop multimodal agents capable of actively engaging and utilizing tools for complex, reasoning-intensive tasks.

\paragraph{Evaluation Benchmarks.}
Current evaluation benchmarks focus on single-round conversations, revealing a gap for multi-round, multi-image conversational benchmarks. 
Such benchmarks require models to respond to a variety of instructions within a single conversation, which could significantly advance research in enhancing models' generalization capabilities and reducing over-fitting to specific instructions.


\bibliographystyle{unsrt}  
\bibliography{references}

\end{document}